%% file: search.tex
\documentclass{article} 
\usepackage{iclr2015,times}
\usepackage{amsmath}
\usepackage{amsthm}
\usepackage{amssymb}
\usepackage{graphicx}
\usepackage{xspace}
\usepackage{tabularx}
\usepackage{url}
\usepackage{natbib}
\usepackage{wrapfig}
\usepackage[utf8]{inputenc}

\usepackage[normalem]{ulem}

\newcommand{\bmx}[0]{\begin{bmatrix}}
\newcommand{\emx}[0]{\end{bmatrix}}

\newcommand{\vect}[1]{\mathbf{#1}}

\newcommand{\vx}[0]{\vect{x}}

\newcommand{\vy}[0]{\vect{y}}

\newcommand{\RR}[0]{\mathbb{R}}

\DeclareMathOperator*{\argmax}{\arg \max}

\newcommand{\ola}{\overleftarrow}
\newcommand{\ora}{\overrightarrow}
\newcommand{\ov}{\overline}

\graphicspath{ {./figures/} }

\title{Neural Machine Translation \\ by Jointly Learning to Align and Translate}

\author{
Dzmitry Bahdanau \\
Jacobs University Bremen, Germany
\And
KyungHyun Cho~ ~ ~ ~Yoshua Bengio\thanks{CIFAR Senior Fellow} \\
Universit\'{e} de Montr\'{e}al
}

\iclrfinalcopy
\iclrconference 
\iclrfinalcopy

\begin{document}

\maketitle

\input{main.tex}

\bibliography{strings,strings-shorter,ml,aigaion,myref}
\bibliographystyle{natbib}

\newpage
\appendix
\input{supp.tex}

\end{document}

%% file: main.tex
\begin{abstract}
    Neural machine translation is a recently proposed approach to machine
    translation. Unlike the traditional statistical machine translation, the
    neural machine translation aims at building a single neural network that
    can be jointly tuned to maximize the translation performance. The models
    proposed recently for neural machine translation often belong to a family
    of encoder--decoders and encode a source sentence into a fixed-length
    vector from which a decoder generates a translation. In this paper, we
    conjecture that the use of a fixed-length vector is a bottleneck in
    improving the performance of this basic encoder--decoder architecture, and
    propose to extend this by allowing a model to automatically
    \mbox{(soft-)search} for parts of a source sentence that are relevant to
    predicting a target word, without having to form these parts as a hard
    segment explicitly. With this new approach, we achieve a translation
    performance comparable to the existing state-of-the-art phrase-based system
    on the task of English-to-French translation. Furthermore, qualitative
    analysis reveals that the \mbox{(soft-)alignments} found by the model agree
    well with our intuition.
\end{abstract}

\section{Introduction}

{\it Neural machine translation} is a newly emerging approach to machine
translation, recently proposed by \citet{Kalchbrenner2013},
\citet{Sutskever2014} and \citet{Cho2014a}. Unlike the traditional phrase-based
translation system~\citep[see, e.g.,][]{Koehn2003} which consists of many small
sub-components that are tuned separately, neural machine translation attempts
to build and train a single, large neural network that reads a sentence and
outputs a correct translation. 

Most of the proposed neural machine translation models belong to a family of
{\it encoder--decoders}~\citep{Sutskever2014,Cho2014}, with an encoder and a
decoder for each language, or involve a language-specific encoder applied to
each sentence whose outputs are then compared~\citep{Hermann2014}.  An encoder
neural network reads and encodes a source sentence into a fixed-length vector.
A decoder then outputs a translation from the encoded vector. The whole
encoder--decoder system, which consists of the encoder and the decoder for a
language pair, is jointly trained to maximize the probability of a correct
translation given a source sentence.

A potential issue with this encoder--decoder approach is that a neural network
needs to be able to compress all the necessary information of a source sentence
into a fixed-length vector.  This may make it difficult for the neural network
to cope with long sentences, especially those that are longer than the
sentences in the training corpus.  \citet{Cho2014a} showed that indeed the
performance of a basic encoder--decoder deteriorates rapidly as the length of
an input sentence increases. 

In order to address this issue, we introduce an extension to the
encoder--decoder model which learns to align and translate jointly. Each time
the proposed model generates a word in a translation, it \mbox{(soft-)searches}
for a set of positions in a source sentence where the most relevant information
is concentrated.  The model then predicts a target word based on the context
vectors associated with these source positions and all the previous generated
target words. 

The most important distinguishing feature of this approach from the basic
encoder--decoder is that it does not attempt to encode a whole input sentence
into a single fixed-length vector. Instead, it encodes the input sentence into
a sequence of vectors and chooses a subset of these vectors adaptively while
decoding the translation. This frees a neural translation model from having to
squash all the information of a source sentence, regardless of its length, into
a fixed-length vector. We show this allows a model to cope better with long
sentences.

In this paper, we show that the proposed approach of jointly learning to align
and translate achieves significantly improved translation performance over the
basic encoder--decoder approach. The improvement is more apparent with longer
sentences, but can be observed with sentences of any length. On the task of
English-to-French translation, the proposed approach achieves, with a single
model, a translation performance comparable, or close, to the conventional
phrase-based system.  Furthermore, qualitative analysis reveals that the
proposed model finds a linguistically plausible \mbox{(soft-)alignment} between
a source sentence and the corresponding target sentence.

\section{Background: Neural Machine Translation}

From a probabilistic perspective, translation is equivalent to finding a target
sentence $\vy$ that maximizes the conditional probability of $\vy$ given a
source sentence $\vx$, i.e., $\argmax_{\vy} p(\vy \mid \vx)$.  In neural
machine translation, we fit a parameterized model to maximize the conditional
probability of sentence pairs using a parallel training corpus. Once the
conditional distribution is learned by a translation model, given a source
sentence a corresponding translation can be generated by searching for the
sentence that maximizes the conditional probability.

Recently, a number of papers have proposed the use of neural networks to
directly learn this conditional distribution~\citep[see,
e.g.,][]{Kalchbrenner2013,Cho2014,Sutskever2014,Cho2014a,Forcada1997}. This
neural machine translation approach typically consists of two components, the
first of which encodes a source sentence $\vx$ and the second decodes to a
target sentence $\vy$. For instance, two recurrent neural networks (RNN) were
used by \citep{Cho2014} and \citep{Sutskever2014} to encode a variable-length
source sentence into a fixed-length vector and to decode the vector into a
variable-length target sentence.

Despite being a quite new approach, neural machine translation has already
shown promising results. \citet{Sutskever2014} reported that the neural machine
translation based on RNNs with long short-term memory (LSTM) units achieves
close to the state-of-the-art performance of the conventional phrase-based
machine translation system on an English-to-French translation task.\footnote{
    We mean by the state-of-the-art performance, the performance of the
    conventional phrase-based system without using any neural network-based
    component.
} 
Adding neural components to existing translation systems, for instance,
to score the phrase pairs in the phrase table~\citep{Cho2014} or to re-rank
candidate translations~\citep{Sutskever2014}, has allowed to
surpass the previous state-of-the-art performance level.

\subsection{RNN Encoder--Decoder}

Here, we describe briefly the underlying framework, called {\it RNN
Encoder--Decoder}, proposed by \citet{Cho2014} and \citet{Sutskever2014} upon
which we build a novel architecture that learns to align and translate
simultaneously.

In the Encoder--Decoder framework, an encoder reads the input sentence, a
sequence of vectors $\vx=\left( x_1, \cdots, x_{T_x} \right)$, into a vector
$c$.\footnote{
    Although most of the previous works~\citep[see,
    e.g.,][]{Cho2014,Sutskever2014,Kalchbrenner2013} used to encode a
    variable-length input sentence into a {\it fixed-length} vector, it is not
    necessary, and even it may be beneficial to have a {\it variable-length}
    vector, as we will show later.
} The most common approach is to use an RNN such that  
\begin{align}
    \label{eq:forward_state}
    h_t = f\left( x_{t}, h_{t-1} \right)
\end{align}
and
\begin{align*}
    c = q\left(\left\{ h_1, \cdots, h_{T_x} \right\}\right),
\end{align*}
where $h_t \in \RR^{n}$ is a hidden state at time $t$, and $c$ is a vector
generated from the sequence of the hidden states. $f$ and $q$ are some
nonlinear functions. \citet{Sutskever2014} used an LSTM as $f$ and
$q\left(\left\{ h_1, \cdots, h_T \right\}\right)=h_T$, for instance.

The decoder is often trained to predict the next word $y_{t'}$ given the context
vector $c$ and all the previously predicted words $\left\{ y_1, \cdots, y_{t'-1}
\right\}$. In other words, the decoder defines a probability over the
translation $\vy$ by decomposing the joint probability into the ordered
conditionals:
\begin{align}
    \label{eq:decoder_prob}
    p(\vy) = \prod_{t=1}^T p(y_t \mid \left\{ y_1, \cdots, y_{t-1} \right\}, c),
\end{align}
where $\vy = \left( y_1, \cdots, y_{T_y} \right)$. With an RNN, each
conditional probability is modeled as
\begin{align}
    \label{eq:output_rnn}
    p(y_t \mid \left\{ y_1, \cdots, y_{t-1} \right\}, c) = g(y_{t-1}, s_{t}, c),
\end{align}
where $g$ is a nonlinear, potentially multi-layered, function that outputs the
probability of $y_t$, and $s_t$ is the hidden state of the RNN.  It should be
noted that other architectures such as a hybrid of an RNN and a
de-convolutional neural network can be used~\citep{Kalchbrenner2013}.

\section{Learning to Align and Translate}
\label{sec:main}

In this section, we propose a novel architecture for neural machine
translation.  The new architecture consists of a bidirectional RNN as an
encoder (Sec.~\ref{sec:birnn_encoder}) and a decoder that emulates
searching through a source sentence during decoding a translation
(Sec.~\ref{sec:search_decoder}). 

\subsection{Decoder: General Description}
\label{sec:search_decoder}

\begin{wrapfigure}{R}{0.3\textwidth}
    \centering
    \includegraphics[width=0.29\textwidth]{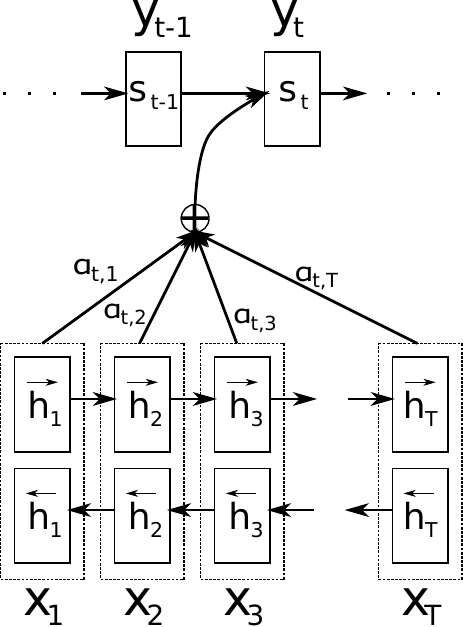}
    \caption{
        The graphical illustration of the proposed model trying to
        generate the $t$-th target word $y_t$ given a source sentence $(x_1, x_2,
        \dots, x_T)$.
    }
    \label{fig:rnnsearch}
\end{wrapfigure}

In a new model architecture, we define each conditional probability in
Eq.~\eqref{eq:decoder_prob} as:
\begin{align}
    \label{eq:generate_y}
    p(y_{i}|y_1, \ldots, y_{i-1}, \vx) = g(y_{i-1}, s_{i}, c_i),
\end{align}
where $s_{i}$ is an RNN hidden state for time $i$, computed by
\[
    s_i = f(s_{i-1}, y_{i-1}, c_{i}).
\]
It should be noted that unlike the existing encoder--decoder approach (see
Eq.~\eqref{eq:decoder_prob}), here the probability is conditioned on a distinct
context vector $c_i$ for each target word $y_i$. 

The context vector $c_i$ depends on a sequence of {\em annotations} $(h_1,
\cdots, h_{T_x})$ to which an encoder maps the input sentence.  Each annotation
$h_i$ contains information about the whole input sequence with a strong focus
on the parts surrounding the $i$-th word of the input sequence. We explain in
detail how the annotations are computed in the next section.

The context vector $c_i$ is, then, computed as a weighted sum of these
annotations $h_i$:
\begin{align}
    \label{eq:context_vector}
    c_i = \sum_{j=1}^{T_x} \alpha_{ij} h_j.
\end{align}
The weight $\alpha_{ij}$ of each annotation $h_j$ is computed by 
\begin{align}
    \label{eq:annotation_weight}
    \alpha_{ij} = \frac{\exp\left(e_{ij}\right)}{\sum_{k=1}^{T_x} \exp\left(e_{ik}\right)},
\end{align}
where
\[
    e_{ij} = a(s_{i-1}, h_j)
\]
is an {\it alignment model} which scores how well the inputs around
position $j$ and the output at position $i$ match. The score is based on
the RNN hidden state $s_{i-1}$ (just before emitting $y_i$,
Eq.~\eqref{eq:generate_y}) and the $j$-th annotation $h_j$ of the input
sentence.

We parametrize the alignment model $a$ as a feedforward neural network which is
jointly trained with all the other components of the proposed system.  Note
that unlike in traditional machine translation, the alignment is not considered
to be a latent variable. Instead, the alignment model directly computes a soft
alignment, which allows the gradient of the cost function to be backpropagated
through. This gradient can be used to train the alignment model as well as the
whole translation model jointly.

We can understand the approach of taking a weighted sum of all the annotations
as computing an {\em expected annotation}, where the expectation is over
possible alignments.  Let $\alpha_{ij}$ be a probability that the target word
$y_i$ is aligned to, or translated from, a source word $x_j$. Then, the $i$-th
context vector $c_i$ is the expected annotation over all the annotations with
probabilities $\alpha_{ij}$.

The probability $\alpha_{ij}$, or its associated energy $e_{ij}$, reflects the
importance of the annotation $h_j$ with respect to the previous hidden state
$s_{i-1}$ in deciding the next state $s_i$ and generating $y_i$.  Intuitively,
this implements a mechanism of attention in the decoder. The decoder decides
parts of the source sentence to pay attention to. By letting the decoder have an
attention mechanism, we relieve the encoder from the burden of having to encode
all information in the source sentence into a fixed-length vector. With this new
approach the information can be spread throughout the sequence of annotations,
which can be selectively retrieved by the decoder accordingly.

\subsection{Encoder: Bidirectional RNN for Annotating Sequences}
\label{sec:birnn_encoder}

The usual RNN, described in Eq.~\eqref{eq:forward_state}, reads an input
sequence $\vx$ in order starting from the first symbol $x_1$ to the last one
$x_{T_x}$. However, in the proposed scheme, we would like the annotation of each
word to summarize not only the preceding words, but also the following words.
Hence, we propose to use a bidirectional RNN~\citep[BiRNN, ][]{Schuster1997},
which has been successfully used recently in speech recognition~\citep[see,
e.g.,][]{Graves2013asru}. 

A BiRNN consists of forward and backward RNN's. The forward RNN $\ora{f}$ reads
the input sequence as it is ordered (from $x_1$ to $x_{T_x}$) and calculates a
sequence of {\it forward hidden states} $( \ora{h}_1, \cdots, \ora{h}_{T_x})$.
The backward RNN $\ola{f}$ reads the sequence in the reverse order (from
$x_{T_x}$ to $x_1$), resulting in a sequence of {\it backward hidden states} $(
\ola{h}_1, \cdots, \ola{h}_{T_x})$. 

We obtain an annotation for each word $x_j$ by concatenating the forward hidden
state $\ora{h}_j$ and the backward one $\ola{h}_j$, i.e., $h_j = \left[
\ora{h}_j^\top ; \ola{h}_j^\top \right]^\top$. In this way, the annotation $h_j$
contains the summaries of both the preceding words and the following words. Due
to the tendency of RNNs to better represent recent inputs, the annotation $h_j$
will be focused on the words around $x_j$. This sequence of annotations is used
by the decoder and the alignment model later to compute the context vector
(Eqs.~\eqref{eq:context_vector}--\eqref{eq:annotation_weight}).

See Fig.~\ref{fig:rnnsearch} for the graphical illustration of the proposed
model.

\section{Experiment Settings}
\label{sec:exp_settings}

We evaluate the proposed approach on the task of English-to-French translation.
We use the bilingual, parallel corpora provided by ACL WMT '14.\footnote{
    \url{http://www.statmt.org/wmt14/translation-task.html}
} As a comparison, we also report the performance of an RNN Encoder--Decoder
which was proposed recently by \citet{Cho2014}. We use the same training
procedures and the same dataset for both models.\footnote{
    Implementations are available at \url{https://github.com/lisa-groundhog/GroundHog}.
}

\begin{figure}[t]
    \centering
    \begin{minipage}{0.7\textwidth}
        \includegraphics[width=\textwidth,clip=True,trim=15 20 15 20]{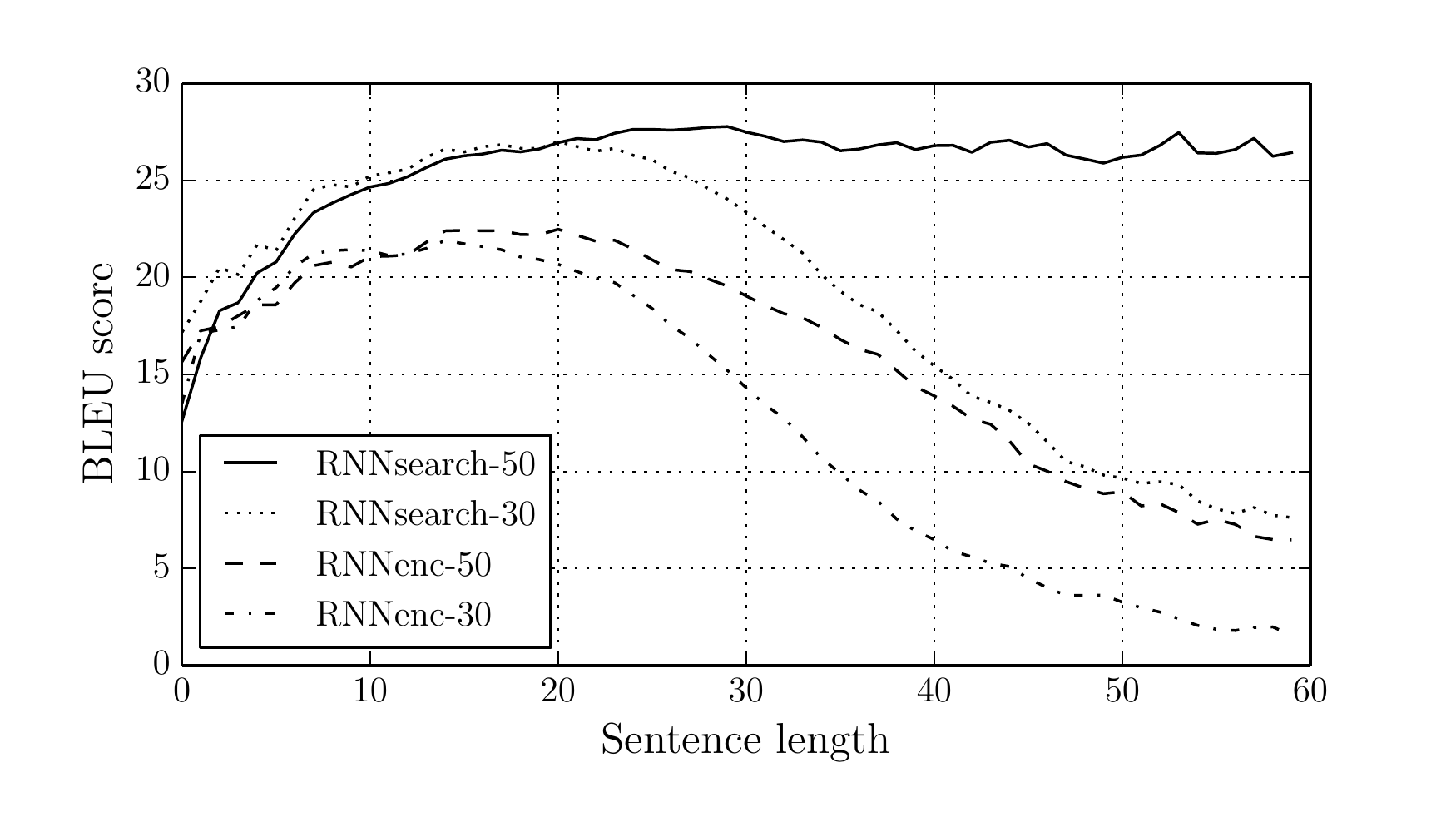}
    \end{minipage}
    \hfill
    \begin{minipage}[t!]{0.29\textwidth}
        \caption{
            The BLEU scores of the generated translations on the test set with
            respect to the lengths of the sentences. The results are on the full
            test set which includes sentences having unknown words to the
            models.
        }
        \label{fig:bleu_length}
    \end{minipage}
\end{figure}

\subsection{Dataset}

WMT '14 contains the following English-French parallel corpora: Europarl (61M
words), news commentary (5.5M), UN (421M) and two crawled corpora of 90M and
272.5M words respectively, totaling 850M words. 
Following the procedure described
in \citet{Cho2014}, we reduce the size of the combined corpus to have 348M words
using the data selection method by \citet{Axelrod2011}.\footnote{
    Available online at
    \url{http://www-lium.univ-lemans.fr/~schwenk/cslm_joint_paper/}.
} We do not use any monolingual data other than the mentioned parallel corpora,
although it may be possible to use a much larger monolingual corpus to pretrain
an encoder. We concatenate news-test-2012 and news-test-2013 to make a
development (validation) set, and evaluate the models on the test set
(news-test-2014) from WMT '14, which consists of 3003 sentences not present in
the training data. 

After a usual tokenization\footnote{
    We used the tokenization script from the open-source machine translation
    package, Moses. 
}, 
we use a shortlist of 30,000 most frequent words in each language to train our
models. Any word not included in the shortlist is mapped to a special token
($\left[ \mbox{UNK} \right]$). We do not apply any other special preprocessing,
such as lowercasing or stemming, to the data.

\begin{figure}[t]
    \centering
    \begin{minipage}[b]{0.48\textwidth}
        \raggedleft
        \includegraphics[width=1.\columnwidth]{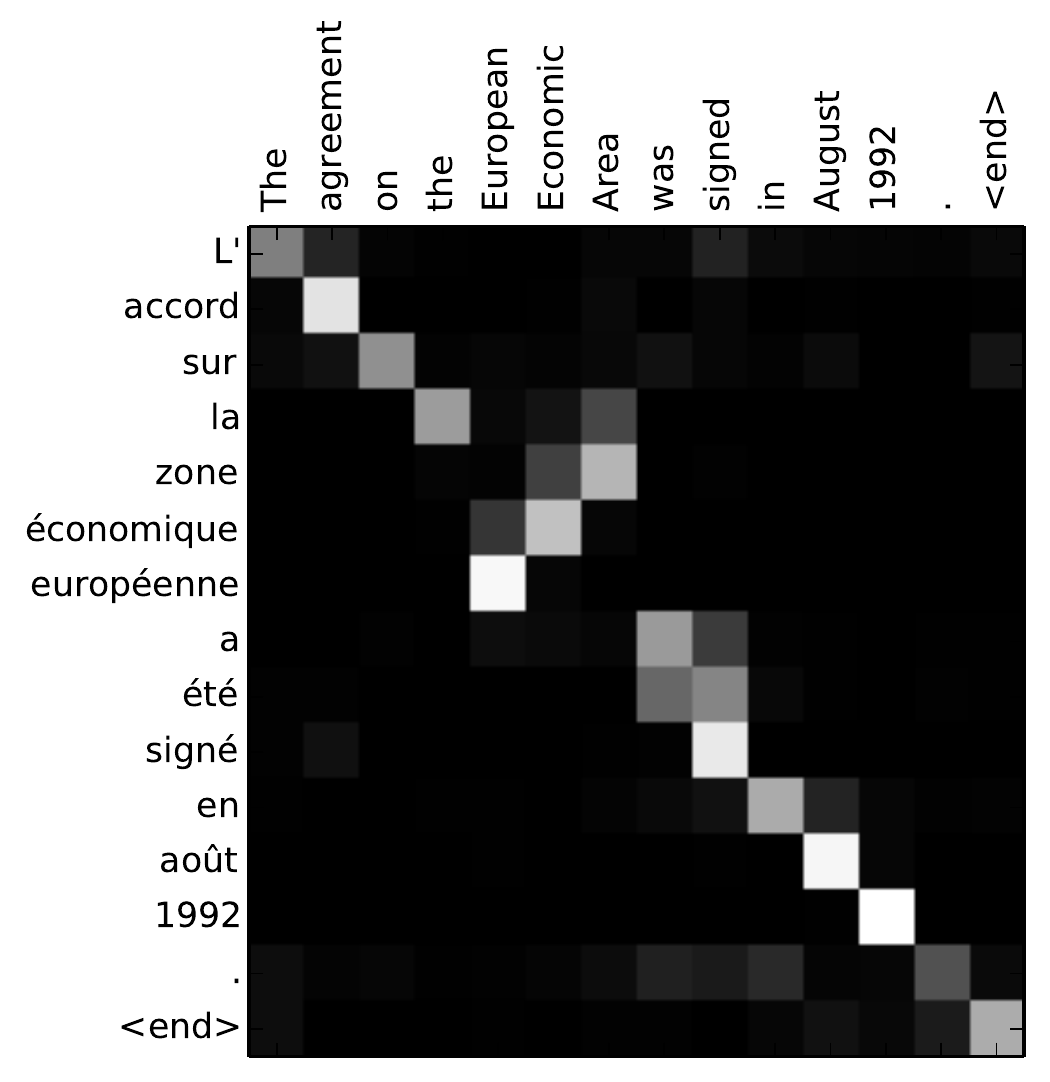}
    \end{minipage}
    \hfill
    \begin{minipage}[b]{0.48\textwidth}
        \raggedleft
        \includegraphics[width=1.\columnwidth]{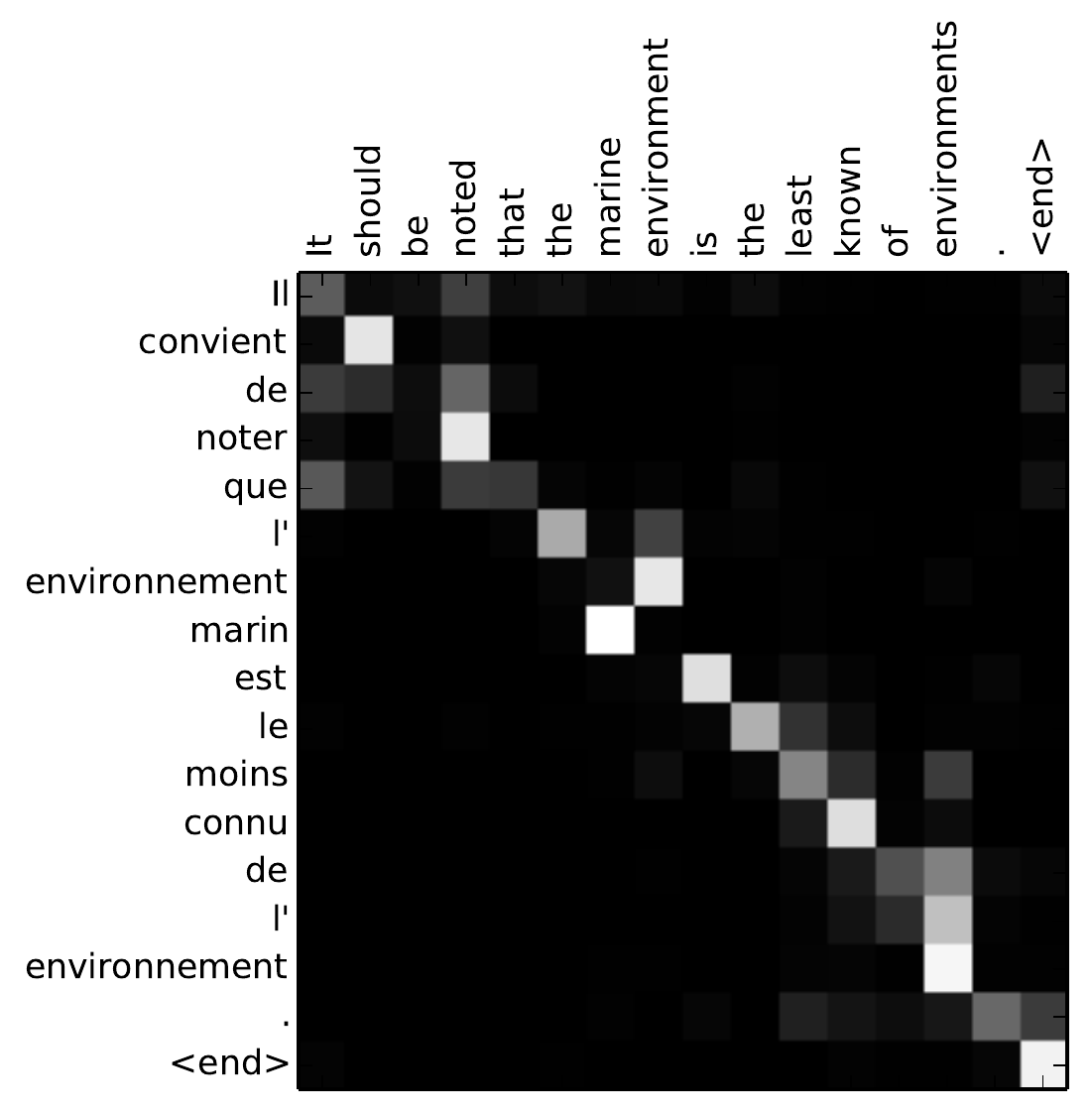}
    \end{minipage}

    \begin{minipage}{0.48\textwidth}
        \centering
        (a)
    \end{minipage}
    \hfill
    \begin{minipage}{0.48\textwidth}
        \centering
        (b)
    \end{minipage}

    \begin{minipage}[b]{0.48\textwidth}
        \raggedleft
        \includegraphics[width=1.\columnwidth]{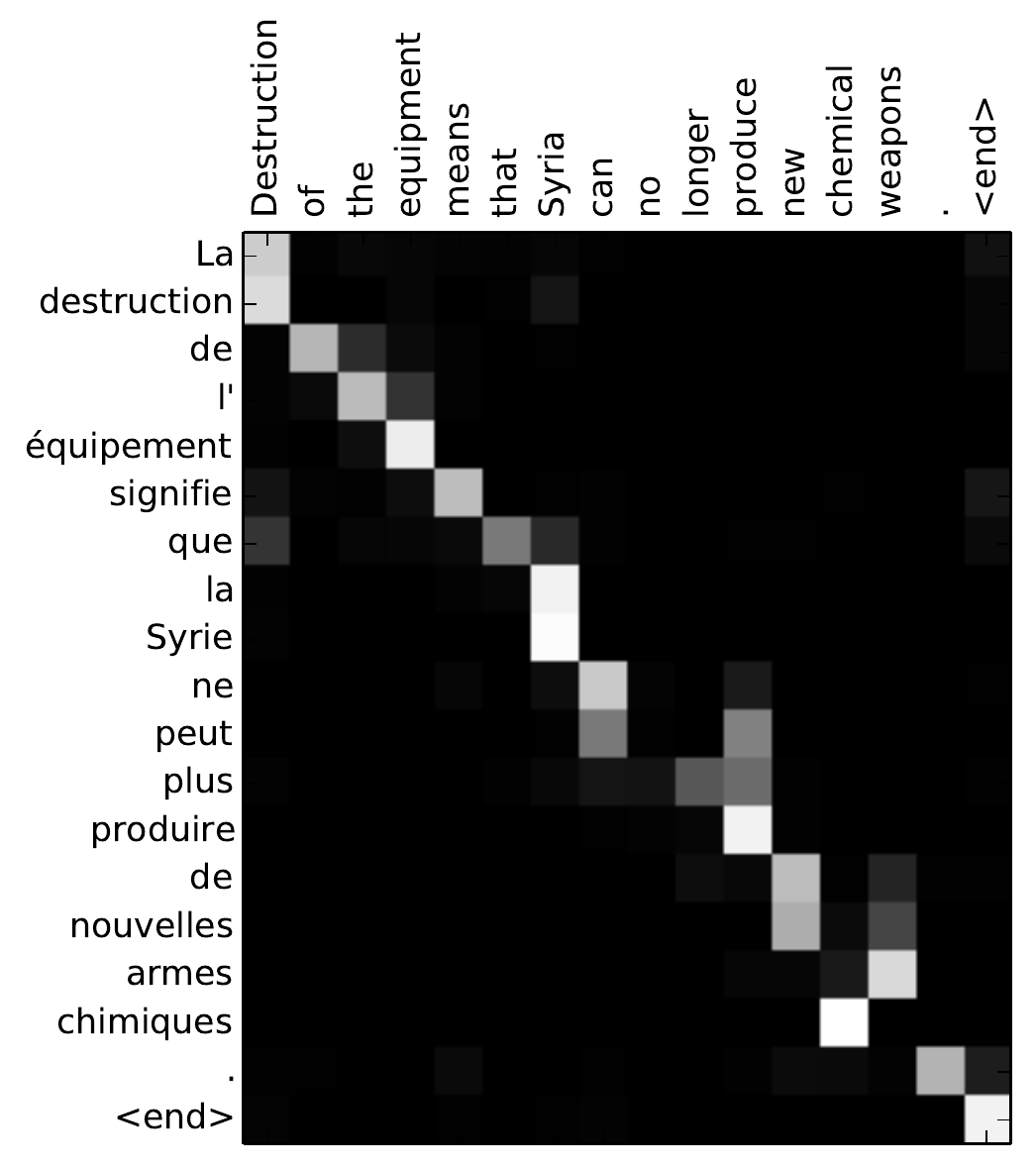}
    \end{minipage}
    \hfill
    \begin{minipage}[b]{0.48\textwidth}
        \raggedleft
        \includegraphics[width=1.\columnwidth]{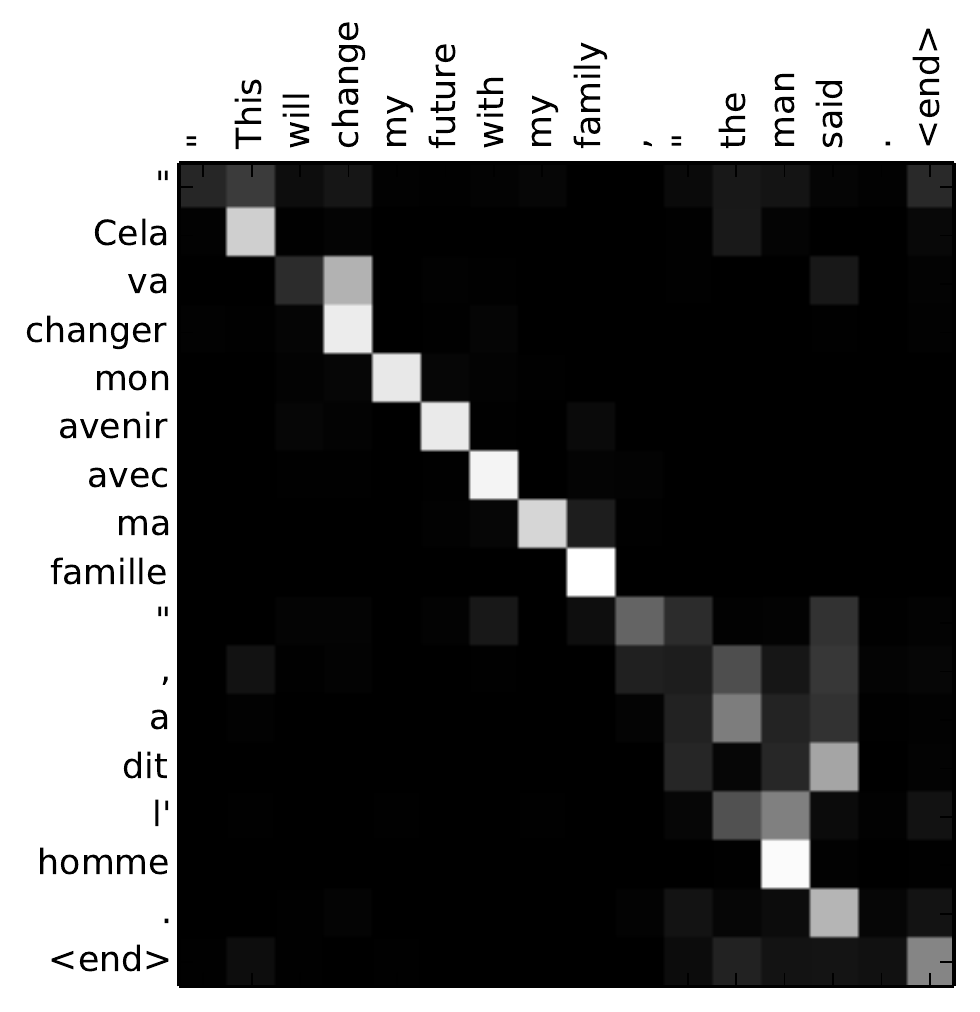}
    \end{minipage}

    \begin{minipage}{0.48\textwidth}
        \centering
        (c)
    \end{minipage}
    \hfill
    \begin{minipage}{0.48\textwidth}
        \centering
        (d)
    \end{minipage}

    \caption{
        Four sample alignments found by RNNsearch-50. The x-axis and y-axis of
        each plot correspond to the words in the source sentence (English) and
        the generated translation (French), respectively. Each pixel shows the
        weight $\alpha_{ij}$ of the annotation of the $j$-th source word for the
        $i$-th target word (see Eq.~\eqref{eq:annotation_weight}), in grayscale
        ($0$: black, $1$: white). (a) an arbitrary sentence. (b--d) three
        randomly selected samples among the sentences without any unknown words
        and of length between 10 and 20 words from the test set.
    }
    \label{fig:alignment}
\end{figure}

\subsection{Models}

We train two types of models. The first one is an RNN
Encoder--Decoder~\citep[RNNencdec,][]{Cho2014}, and the other is the proposed
model, to which we refer as RNNsearch. We train each model twice: first with the
sentences of length up to 30 words (RNNencdec-30, RNNsearch-30) and then with the
sentences of length up to 50 word (RNNencdec-50, RNNsearch-50).

The encoder and decoder of the RNNencdec have 1000 hidden units each.\footnote{
    In this paper, by a 'hidden unit', we always mean the gated hidden unit (see
    Appendix~\ref{sec:gatedrnn}).
} The encoder of the RNNsearch consists of forward and backward recurrent neural
networks (RNN) each having 1000 hidden units. Its decoder has 1000 hidden units.
In both cases, we use a multilayer network with a single
maxout~\citep{Goodfellow2013} hidden layer to compute the conditional
probability of each target word~\citep{Pascanu2014rec}.

We use a minibatch stochastic gradient descent (SGD) algorithm together with
Adadelta~\citep{Zeiler2012} to train each model. Each SGD update direction is
computed using a minibatch of 80 sentences. We trained each model for
approximately 5 days.

Once a model is trained, we use a beam search to find a translation that
approximately maximizes the conditional probability~\citep[see,
e.g.,][]{Graves2012,Boulanger2013}. \citet{Sutskever2014} used this approach to
generate translations from their neural machine translation model.

For more details on the architectures of the models and training procedure used
in the experiments, see Appendices~\ref{sec:model_detail} and
\ref{sec:training_detail}.

\section{Results}
\label{sec:exp_results}

\subsection{Quantitative Results}

\begin{table}[t]
    \centering
    \hfill
    \begin{minipage}{0.45\textwidth}
        \begin{tabular}{c|c|c}
        Model & All & No UNK$^\circ$ \\
        \hline
        \hline
        RNNencdec-30 & 13.93 & 24.19 \\
        RNNsearch-30 & 21.50 & 31.44 \\
        \hline
        RNNencdec-50 & 17.82 & 26.71 \\
        RNNsearch-50 & 26.75 & 34.16 \\
        \hline
        RNNsearch-50$^\star$ & 28.45 & 36.15 \\
        \hline
        Moses & 33.30 & 35.63
        \end{tabular}
    \end{minipage}
    \begin{minipage}[t!]{0.53\textwidth}
        \vspace{-2mm}

        \caption{BLEU scores of the trained models computed on the test set.
            The second and third columns show respectively the scores on all the
            sentences and, on the sentences without any unknown word in
            themselves and in the reference translations. Note that
            RNNsearch-50$^\star$ was trained much longer until the performance
            on the development set stopped improving.  ($\circ$) We disallowed
            the models to generate [UNK] tokens when only the sentences having
            no unknown words were evaluated (last column).  
        }
        \label{tab:bleu}
    \end{minipage}
    \hfill
\end{table}

In Table~\ref{tab:bleu}, we list the translation performances measured in BLEU
score. It is clear from the table that in all the cases, the proposed RNNsearch
outperforms the conventional RNNencdec.  More importantly, the performance of the
RNNsearch is as high as that of the conventional phrase-based translation system
(Moses), when only the sentences consisting of known words are considered. This
is a significant achievement, considering that Moses uses a separate monolingual
corpus (418M words) in addition to the parallel corpora we used to train the
RNNsearch and RNNencdec.

One of the motivations behind the proposed approach was the use of a
fixed-length context vector in the basic encoder--decoder approach. We
conjectured that this limitation may make the basic encoder--decoder approach to
underperform with long sentences. In Fig.~\ref{fig:bleu_length}, we see that the
performance of RNNencdec dramatically drops as the length of the sentences
increases. On the other hand, both RNNsearch-30 and RNNsearch-50 are more robust
to the length of the sentences. RNNsearch-50, especially, shows no performance
deterioration even with sentences of length 50 or more. This superiority of the
proposed model over the basic encoder--decoder is further confirmed by the fact
that the RNNsearch-30 even outperforms RNNencdec-50 (see Table~\ref{tab:bleu}).

\subsection{Qualitative Analysis}

\subsubsection{Alignment}

The proposed approach provides an intuitive way to inspect the
\mbox{(soft-)alignment} between the words in a generated translation and those
in a source sentence.  This is done by visualizing the annotation weights
$\alpha_{ij}$ from Eq.~\eqref{eq:annotation_weight}, as in
Fig.~\ref{fig:alignment}. Each row of a matrix in each plot indicates the
weights associated with the annotations. From this we see which positions in the
source sentence were considered more important when generating the target word. 

We can see from the alignments in Fig.~\ref{fig:alignment} that the alignment of
words between English and French is largely monotonic. We see strong weights
along the diagonal of each matrix. However, we also observe a number of
non-trivial, non-monotonic alignments. Adjectives and nouns are typically
ordered differently between French and English, and we see an example in
Fig.~\ref{fig:alignment}~(a). From this figure, we see that the model correctly
translates a phrase [European Economic Area] into [zone \'economique
europ\'een]. The RNNsearch was able to correctly align [zone] with [Area],
jumping over the two words ([European] and [Economic]), and then looked one word
back at a time to complete the whole phrase [zone \'economique europ\'eenne]. 

The strength of the soft-alignment, opposed to a hard-alignment, is evident, for
instance, from Fig.~\ref{fig:alignment}~(d). Consider the source phrase [the
man] which was translated into [l' homme]. Any hard alignment will map [the] to
[l'] and [man] to [homme]. This is not helpful for translation, as one must
consider the word following [the] to determine whether it should be translated
into [le], [la], [les] or [l'].  Our soft-alignment solves this issue naturally
by letting the model look at both [the] and [man], and in this example, we see
that the model was able to correctly translate [the] into [l']. We observe
similar behaviors in all the presented cases in Fig.~\ref{fig:alignment}. An
additional benefit of the soft alignment is that it naturally deals with source
and target phrases of different lengths, without requiring a counter-intuitive
way of mapping some words to or from nowhere ([NULL])~\citep[see, e.g.,
Chapters~4 and 5 of][]{Koehn2010}.

\subsubsection{Long Sentences}

As clearly visible from Fig.~\ref{fig:bleu_length} the proposed model
(RNNsearch) is much better than the conventional model (RNNencdec) at translating
long sentences. This is likely due to the fact that the RNNsearch does not
require encoding a long sentence into a fixed-length vector perfectly, but only
accurately encoding the parts of the input sentence that surround a particular
word.

As an example, consider this source sentence from the test set:
\begin{quote}
    \it
An admitting privilege is the right of a doctor to admit a patient to a hospital
or a medical centre \uline{to carry out a diagnosis or a procedure, based on
his status as a health care worker at a hospital.}
\end{quote}
The RNNencdec-50 translated this sentence into:
\begin{quote}
    \it
Un privilège d'admission est le droit d'un m\'edecin de reconnaître un patient à
l'h\^opital ou un centre m\'edical \uline{d'un diagnostic ou de prendre un diagnostic
en fonction de son \'etat de sant\'e}.
\end{quote}

The RNNencdec-50 correctly translated the source sentence until [a medical center].
However, from there on (underlined), it deviated from the original meaning of
the source sentence. For instance, it replaced [based on his status as a health
care worker at a hospital] in the source sentence with [en fonction de son
\'etat de sant\'e] (``based on his state of health'').

On the other hand, the RNNsearch-50 generated the following correct translation,
preserving the whole meaning of the input sentence without omitting any details:
\begin{quote}
{    \it
Un privilège d'admission est le droit d'un m\'edecin d'admettre un patient à un
h\^opital ou un centre m\'edical \uline{pour effectuer un diagnostic ou une
proc\'edure, selon son statut de travailleur des soins de sant\'e à
l'h\^opital.}}
\end{quote}

Let us consider another sentence from the test set:
\begin{quote}
{    \it
This kind of experience is part of Disney's efforts to "extend the lifetime of
its series and build new relationships with audiences \uline{via digital platforms that
are becoming ever more important," he added.} }
\end{quote}
The translation by the RNNencdec-50 is
\begin{quote}
{    \it
Ce type d'exp\'erience fait partie des initiatives du Disney pour "prolonger la
dur\'ee de vie de ses nouvelles et de d\'evelopper des liens avec les
\uline{lecteurs num\'eriques qui deviennent plus complexes.} }
\end{quote}

As with the previous example, the RNNencdec began deviating from the actual meaning
of the source sentence after generating approximately 30 words (see the
underlined phrase). After that point, the quality of the translation
deteriorates, with basic mistakes such as the lack of a closing quotation mark.

Again, the RNNsearch-50 was able to translate this long sentence correctly:
\begin{quote}
{    \it
Ce genre d'exp\'erience fait partie des efforts de Disney pour "prolonger la
dur\'ee de vie de ses s\'eries et cr\'eer de nouvelles relations avec des
publics \uline{via des plateformes num\'eriques de plus en plus importantes", a-t-il
ajout\'e.} }
\end{quote}

In conjunction with the quantitative results presented already, these
qualitative observations confirm our hypotheses that the RNNsearch architecture
enables far more reliable translation of long sentences than the standard RNNencdec
model. 

In Appendix~\ref{sec:long_translation}, we provide a few more sample
translations of long source sentences generated by the RNNencdec-50, RNNsearch-50
and Google Translate along with the reference translations.

\section{Related Work}

\subsection{Learning to Align}

A similar approach of aligning an output symbol with an input symbol was
proposed recently by \citet{Graves2013} in the context of handwriting synthesis.
Handwriting synthesis is a task where the model is asked to generate handwriting
of a given sequence of characters. In his work, he used a mixture of Gaussian
kernels to compute the weights of the annotations, where the location, width and
mixture coefficient of each kernel was predicted from an alignment model. More
specifically, his alignment was restricted to predict the location such that the
location increases monotonically.

The main difference from our approach is that, in \citep{Graves2013}, the modes
of the weights of the annotations only move in one direction. In the context of
machine translation, this is a severe limitation, as (long-distance) reordering
is often needed to generate a grammatically correct translation (for instance,
English-to-German).

Our approach, on the other hand, requires computing the annotation weight of
every word in the source sentence for each word in the translation. This
drawback is not severe with the task of translation in which most of input and
output sentences are only 15--40 words. However, this may limit the
applicability of the proposed scheme to other tasks.

\subsection{Neural Networks for Machine Translation}

Since \citet{Bengio2003lm} introduced a neural probabilistic language model
which uses a neural network to model the conditional probability of a word given
a fixed number of the preceding words, neural networks have widely been used in
machine translation. However, the role of neural networks has been largely
limited to simply providing a single feature to an existing statistical machine
translation system or to re-rank a list of candidate translations provided by an
existing system. 

For instance, \citet{Schwenk2012} proposed using a feedforward neural network to
compute the score of a pair of source and target phrases and to use the score as
an additional feature in the phrase-based statistical machine translation
system. More recently, \citet{Kalchbrenner2013} and \citet{Devlin2014} reported
the successful use of the neural networks as a sub-component of the existing
translation system. Traditionally, a neural network trained as a target-side
language model has been used to rescore or rerank a list of candidate
translations~\citep[see, e.g.,][]{Schwenk2006t}.

Although the above approaches were shown to improve the translation performance
over the state-of-the-art machine translation systems, we are more interested in
a more ambitious objective of designing a completely new translation system
based on neural networks. The neural machine translation approach we consider in
this paper is therefore a radical departure from these earlier works. Rather
than using a neural network as a part of the existing system, our model works on
its own and generates a translation from a source sentence directly.

\section{Conclusion}

The conventional approach to neural machine translation, called an
encoder--decoder approach, encodes a whole input sentence into a fixed-length
vector from which a translation will be decoded. We conjectured that the use of
a fixed-length context vector is problematic for translating long sentences,
based on a recent empirical study reported by \citet{Cho2014a} and
\citet{Pouget2014}.

In this paper, we proposed a novel architecture that addresses this issue. We
extended the basic encoder--decoder by letting a model \mbox{(soft-)search} for
a set of input words, or their annotations computed by an encoder, when
generating each target word. This frees the model from having to encode a whole
source sentence into a fixed-length vector, and also lets the model focus only
on information relevant to the generation of the next target word. This has a
major positive impact on the ability of the neural machine translation system to
yield good results on longer sentences. Unlike with the traditional machine
translation systems, all of the pieces of the translation system, including the
alignment mechanism, are jointly trained towards a better log-probability of
producing correct translations.

We tested the proposed model, called RNNsearch, on the task of English-to-French
translation. The experiment revealed that the proposed RNNsearch outperforms the
conventional encoder--decoder model (RNNencdec) significantly, regardless of the
sentence length and that it is much more robust to the length of a source
sentence. From the qualitative analysis where we investigated the
\mbox{(soft-)alignment} generated by the RNNsearch, we were able to conclude
that the model can correctly align each target word with the relevant words, or
their annotations, in the source sentence as it generated a correct translation.

Perhaps more importantly, the proposed approach achieved a translation
performance comparable to the existing phrase-based statistical machine
translation. It is a striking result, considering that the proposed
architecture, or the whole family of neural machine translation, has only been
proposed as recently as this year. We believe the architecture proposed here is
a promising step toward better machine translation and a better understanding of
natural languages in general.

One of challenges left for the future is to better handle unknown, or rare
words. This will be required for the model to be more widely used and to match
the performance of current state-of-the-art machine translation systems in all
contexts.

\section*{Acknowledgments}

The authors would like to thank the developers of
Theano~\citep{bergstra+al:2010-scipy,Bastien-Theano-2012}.  We  acknowledge the
support of the following agencies for research funding and computing support:
NSERC, Calcul Qu\'{e}bec, Compute Canada, the Canada Research Chairs and CIFAR.
Bahdanau thanks the support from Planet Intelligent Systems GmbH.  We also thank
Felix Hill, Bart van Merri\'enboer, Jean Pouget-Abadie, Coline Devin and
Tae-Ho Kim.

%% file: supp.tex
\section{Model Architecture}
\label{sec:model_detail}

\subsection{Architectural Choices}

The proposed scheme in Section~\ref{sec:main} is a general framework where one
can freely define, for instance, the activation functions $f$ of recurrent
neural networks (RNN) and the alignment model $a$. Here, we describe the
choices we made for the experiments in this paper. 

\subsubsection{Recurrent Neural Network}
\label{sec:gatedrnn}

For the activation function $f$ of an RNN, we use the gated hidden unit
recently proposed by \citet{Cho2014}. The gated hidden unit is an alternative
to the conventional {\it simple} units such as an element-wise $\tanh$.  This
gated unit is similar to a long short-term memory (LSTM) unit proposed earlier
by \citet{Hochreiter+Schmidhuber-1997}, sharing with it the ability to better
model and learn long-term dependencies. This is made possible by having
computation paths in the unfolded RNN for which the product of derivatives is
close to 1.  These paths allow gradients to flow backward easily without
suffering too much from the vanishing
effect~\citep{Hochreiter91,Bengio-trnn93,Pascanu+al-ICML2013-small}. It is
therefore possible to use LSTM units instead of the gated hidden unit described
here, as was done in a similar context by \citet{Sutskever2014}.

The new state $s_i$ of the RNN employing $n$ gated hidden units\footnote{
    Here, we show the formula of the decoder. The same formula can be used in
    the encoder by simply ignoring the context vector $c_i$ and the related
    terms.
}
is computed by
\begin{align*}
    s_i = f(s_{i-1}, y_{i-1}, c_i) = (1 - z_i) \circ s_{i-1} + z_i \circ \tilde{s}_{i},
\end{align*}
where $\circ$ is an element-wise multiplication, and $z_i$ is the output of the
update gates (see below). The proposed updated state $\tilde{s}_{i}$ is computed
by
\begin{align*}
    \tilde{s}_{i} = \tanh \left( W e(y_{i - 1}) + U \left[ r_i \circ s_{i - 1} \right] +
    C c_i \right),
\end{align*}
where $e(y_{i-1}) \in \RR^{m}$ is an $m$-dimensional embedding of a word
$y_{i-1}$, and $r_i$ is the output of the reset gates (see below).  When $y_i$
is represented as a $1$-of-$K$ vector, $e(y_i)$ is simply a column of an
embedding matrix $E \in \RR^{m \times K}$. Whenever possible, we omit bias terms
to make the equations less cluttered.

The update gates $z_i$ allow each hidden unit to maintain its previous
activation, and the reset gates $r_i$ control how much and what information from
the previous state should be reset. We compute them by
\begin{align*}
    z_i &= \sigma \left( W_{z} e(y_{i - 1}) + U_{z} s_{i - 1} + C_{z} c_i\right), \\
    r_i &= \sigma \left( W_{r} e(y_{i - 1}) + U_{r} s_{i - 1} + C_{r} c_i\right),
\end{align*}
where $\sigma\left(\cdot\right)$ is a logistic sigmoid function.

At each step of the decoder, we compute the output probability
(Eq.~\eqref{eq:generate_y}) as a multi-layered function~\citep{Pascanu2014rec}.
We use a single hidden layer of maxout units~\citep{Goodfellow2013} and
normalize the output probabilities (one for each word) with a softmax function
(see Eq.~\eqref{eq:annotation_weight}).

\subsubsection{Alignment Model}

The alignment model should be designed considering that the model needs to be
evaluated $T_x \times T_y$ times for each sentence pair of lengths $T_x$ and
$T_y$. In order to reduce computation, we use a single-layer multilayer
perceptron such that 
\begin{align*}
    a(s_{i-1}, h_j) = v_a^{\top} \tanh\left( W_a s_{i-1} + U_a h_j \right),
\end{align*}
where $W_a \in \RR^{n\times n}, U_a \in \RR^{n\times 2n}$ and $v_a \in \RR^{n}$
are the weight matrices. Since $U_a h_j$ does not depend on $i$, we can
pre-compute it in advance to minimize the computational cost.

\subsection{Detailed Description of the Model}
\subsubsection{Encoder}

In this section, we describe in detail the architecture of the proposed model
(RNNsearch) used in the experiments (see
Sec.~\ref{sec:exp_settings}--\ref{sec:exp_results}).  From here on, we omit all
bias terms in order to increase readability.

The model takes a source sentence of 1-of-K coded word vectors as input
\[
    \vx = (x_1, \ldots, x_{T_x}),\mbox{ }x_i \in \mathbb{R}^{K_x}
\]
and outputs a translated sentence of 1-of-K coded word vectors
\[
    \vy = (y_1, \ldots, y_{T_y}),\mbox{ }y_i \in \mathbb{R}^{K_y},
\]
where $K_x$ and $K_y$ are the vocabulary sizes of source and target languages,
respectively. $T_x$ and $T_y$ respectively denote the lengths of source and
target sentences.

First, the forward states of the bidirectional recurrent neural network (BiRNN)
are computed:
\begin{align*}
    \ora{h}_i =& 
    \begin{cases}
        (1 - \ora{z}_i) \circ \ora{h}_{i-1}  + \ora{z}_i \circ \ora{\underline{h}}_{i} &\mbox{, if }i > 0 \\
        0 &\mbox{, if }i = 0
    \end{cases}
\end{align*}
where
\begin{align*}
    \ora{\underline{h}}_i =& \tanh \left( \ora{W} \ov{E} x_i + \ora{U}\left[ \ora{r}_i \circ \ora{h}_{i-1} \right] \right) \\
    \ora{z}_i =& \sigma\left( \ora{W}_z \ov{E} x_i + \ora{U}_z \ora{h}_{i-1} \right) \\
    \ora{r}_i =& \sigma\left( \ora{W}_r \ov{E} x_i + \ora{U}_r \ora{h}_{i-1} \right).
\end{align*}
$\overline{E} \in \mathbb{R}^{m\times K_x}$ is the word embedding matrix.
$\ora{W}, \ora{W}_z, \ora{W}_r \in \mathbb{R}^{n\times m}$, $\ora{U}, \ora{U}_z,
\ora{U}_r \in \mathbb{R}^{n\times n}$ are weight matrices. $m$ and $n$ are the word
embedding dimensionality and the number of hidden units, respectively. 
$\sigma(\cdot)$ is as usual a logistic sigmoid function.

The backward states $(\ola{h}_1, \cdots, \ola{h}_{T_x})$ are computed similarly.
We share the word embedding matrix $\ov{E}$ between the forward and backward
RNNs, unlike the weight matrices.

We concatenate the forward and backward states to to obtain the annotations
$(h_1, h_2, \cdots, h_{T_x})$, where
\begin{align}
    \label{eq:annotation}
    h_i = \left[ 
        \begin{array}{c}
    \ora{h}_i \\
    \ola{h}_i 
\end{array}
\right]
    \end{align}

\subsubsection{Decoder}

The hidden state $s_i$ of the decoder given the annotations from the encoder is
computed by
\begin{align*}
    s_i =& (1 - z_i) \circ s_{i-1} + z_i \circ \tilde{s}_i,
\end{align*}
where
\begin{align*}
    \tilde{s}_{i} =& \tanh \left( W E y_{i - 1} + U \left[ r_i \circ s_{i - 1} \right] +
    C c_i \right) \\ 
    z_i =& \sigma\left( W_z E y_{i - 1} + U_z s_{i-1} 
    + C_z c_i \right)\\
    r_i =& \sigma\left( W_r E y_{i - 1} + U_r s_{i-1}
    + C_r c_i \right)
\end{align*}
$E$ is the word embedding matrix for the target language.
$W, W_z, W_r \in \mathbb{R}^{n\times m}$, 
$U, U_z, U_r \in \mathbb{R}^{n\times n}$, and
$C, C_z, C_r \in \mathbb{R}^{n\times 2n}$ are weights. Again, $m$ and $n$ are the word
embedding dimensionality and the number of hidden units, respectively.
The initial hidden state $s_0$ is computed by 
$
    s_{0} = \tanh\left( W_s \ola{h}_1 \right),
$
where $W_s \in \RR^{n \times n}$.

The context vector $c_i$ are recomputed at each step by the alignment model:
\begin{align*}
    c_i =& \sum_{j=1}^{T_x} \alpha_{ij} h_j,
\end{align*}
where
\begin{align*}
    \alpha_{ij} =& \frac{\exp\left(e_{ij}\right)}{\sum_{k=1}^{T_x}
    \exp\left(e_{ik}\right)}  \\
    e_{ij} =& v_a^{\top} \tanh\left( W_a s_{i-1} + U_a h_j \right),
\end{align*}
and $h_j$ is the $j$-th annotation in the source sentence (see
Eq.~\eqref{eq:annotation}).  $v_a \in \mathbb{R}^{n'}, W_a \in
\mathbb{R}^{n'\times n}$ and $U_a \in \mathbb{R}^{n'\times  2n}$ are weight
matrices.  Note that the model becomes
RNN Encoder--Decoder~\citep{Cho2014}, if we fix $c_i$ to $\ora{h}_{T_x}$.

With the decoder state $s_{i-1}$, the context $c_{i}$ and the last generated word
$y_{i-1}$, we define the probability of a target word $y_{i}$ as
\begin{align*}
    p(y_{i}|s_i,y_{i-1},c_{i}) \propto& \exp\left(y_{i}^{\top} W_o t_{i}\right),
\end{align*}
where
\begin{align*}
    t_i =&  \left[ \max\left\{\tilde{t}_{i, 2j-1}, \tilde{t}_{i,2j}\right\}
    \right]_{j=1,\ldots,l}^{\top}
\end{align*}
and $\tilde{t}_{i,k}$ is the $k$-th element of a vector $\tilde{t}_i$ which is
computed by
\begin{align*}
    \tilde{t}_{i} =& U_o s_{i - 1} + V_o E y_{i-1} + C_o c_i.
\end{align*}
$W_o \in \mathbb{R}^{K_y\times  l}$, $U_o \in \mathbb{R}^{2l\times n}$, $V_o \in
\mathbb{R}^{2l\times m}$ and $C_o \in \mathbb{R}^{2l\times 2n}$ are weight
matrices. This can be understood as having a deep output~\citep{Pascanu2014rec}
with a single maxout hidden layer~\citep{Goodfellow2013}.

\subsubsection{Model Size}

For all the models used in this paper, the size of a hidden layer $n$ is 1000,
the word embedding dimensionality $m$ is 620 and the size of the maxout hidden
layer in the deep output $l$ is 500. The number of hidden units in the alignment
model $n'$ is 1000.

\begin{table}[t]
    \centering                                                                                         
    \begin{tabular}{c|cccccc}                                                                           
        Model & Updates {\scriptsize ($\times 10^5$)} & Epochs 
        & Hours & GPU & Train NLL & Dev. NLL \\
        \hline                                                           
        \hline
        RNNenc-30 & 8.46 & 6.4 & 109 & {\small TITAN BLACK} & 28.1 & 53.0 \\                                                
        RNNenc-50 & 6.00 & 4.5 & 108 & {\small Quadro K-6000} & 44.0 & 43.6 \\
        \hline
        RNNsearch-30 & 4.71 & 3.6 & 113 & {\small TITAN BLACK} & 26.7 & 47.2 \\
        RNNsearch-50 & 2.88 & 2.2 & 111 & {\small Quadro K-6000} & 40.7 & 38.1 \\
        \hline
    RNNsearch-50$^\star$ & 6.67 & 5.0 & 252 & {\small Quadro K-6000} & 36.7 & 35.2 \\
    \end{tabular}                                                                                      
    \caption{Learning statistics and relevant information. Each update
        corresponds to updating the parameters once using a single minibatch. 
        One epoch is one pass through the training set.
        NLL is the average conditional log-probabilities of the
        sentences in either the training set or the development set. Note that
    the lengths of the sentences differ.}
    \label{tbl:stat}                                                                                   
\end{table}

\section{Training Procedure}
\label{sec:training_detail}

\subsection{Parameter Initialization}

We initialized the recurrent weight matrices $U, U_z, U_r, \ola{U}, \ola{U}_z,
\ola{U}_r, \ora{U}, \ora{U}_z$ and $\ora{U}_r$ as random orthogonal matrices.
For $W_a$ and $U_a$, we initialized them by sampling each element from the
Gaussian distribution of mean $0$ and variance $0.001^2$. All the elements of
$V_a$ and all the bias vectors were initialized to zero. Any other weight matrix
was initialized by sampling from the Gaussian distribution of mean $0$ and
variance $0.01^2$.

\subsection{Training}

We used the stochastic gradient descent (SGD) algorithm.
Adadelta~\citep{Zeiler2012} was used to automatically adapt the learning rate of
each parameter ($\epsilon=10^{-6}$ and $\rho=0.95$). We explicitly normalized
the $L_2$-norm of the gradient of the cost function each time to be at most a
predefined threshold of $1$, when the norm was larger than the
threshold~\citep{Pascanu2013}.  Each SGD update direction was computed with a
minibatch of 80 sentences. 

At each update our implementation requires time proportional to the length of
the longest sentence in a minibatch. Hence, to minimize the waste of
computation, before every 20-th update, we retrieved 1600 sentence pairs, sorted them
according to the lengths and split them into 20 minibatches. The training data
was shuffled once before training and was traversed sequentially in this manner.

In Tables~\ref{tbl:stat} we present the statistics related to training all the
models used in the experiments.

\section{Translations of Long Sentences}
\label{sec:long_translation}

\begin{table}[htp]
    \begin{minipage}{0.99\textwidth}
        \small
        \centering
        \begin{tabular}{p{1.9cm} | p{12cm}}
\hline Source & An admitting privilege is the right of a doctor to admit a patient to a hospital or a medical centre to carry out a diagnosis or a procedure, based on his status as a health care worker at a hospital.
\\
\hline Reference & Le privilège d'admission est le droit d'un m\'edecin, en vertu de son statut de membre soignant d'un h\^opital, d'admettre un patient dans un h\^opital ou un centre m\'edical afin d'y d\'elivrer un diagnostic ou un traitement.
\\
\hline RNNenc-50 & Un privilège d'admission est le droit d'un m\'edecin de reconnaître un patient à l'h\^opital ou un centre m\'edical d'un diagnostic ou de prendre un diagnostic en fonction de son \'etat de sant\'e.
\\
\hline RNNsearch-50 & Un privilège d'admission est le droit d'un m\'edecin d'admettre un patient à un h\^opital ou un centre m\'edical pour effectuer un diagnostic ou une proc\'edure, selon son statut de travailleur des soins de sant\'e à l'h\^opital.
\\
\hline Google \mbox{Translate} & Un privilège admettre est le droit d'un m\'edecin
d'admettre un patient dans un h\^opital ou un centre m\'edical pour effectuer un
diagnostic ou une proc\'edure, fond\'ee sur sa situation en tant que travailleur de
soins de sant\'e dans un h\^opital.
\\
\hline
\multicolumn{2}{c}{} \\
\hline Source & 
This kind of experience is part of Disney's efforts to "extend
the lifetime of its series and build new relationships with audiences via
digital platforms that are becoming ever more important," he added.
\\
\hline Reference & Ce type d'exp\'erience entre dans le cadre des efforts de Disney pour "\'etendre la dur\'ee de vie de ses s\'eries et construire de nouvelles relations avec son public grâce à des plateformes num\'eriques qui sont de plus en plus importantes", a-t-il ajout\'e.
\\
\hline RNNenc-50 & Ce type d'exp\'erience fait partie des initiatives du Disney pour "prolonger la dur\'ee de vie de ses nouvelles et de d\'evelopper des liens avec les lecteurs num\'eriques qui deviennent plus complexes.
\\
\hline RNNsearch-50 & Ce genre d'exp\'erience fait partie des efforts de Disney pour "prolonger la dur\'ee de vie de ses s\'eries et cr\'eer de nouvelles relations avec des publics via des plateformes num\'eriques de plus en plus importantes", a-t-il ajout\'e.
\\
\hline Google \mbox{Translate} & Ce genre d'exp\'erience fait partie des efforts de Disney
à ``\'etendre la dur\'ee de vie de sa s\'erie et construire de nouvelles relations avec
le public par le biais des plates-formes num\'eriques qui deviennent de plus en
plus important'', at-il ajout\'e.
\\
\hline
\multicolumn{2}{c}{} \\
\hline Source & 
In a press conference on Thursday, Mr Blair stated that there was nothing in
this video that might constitute a "reasonable motive" that could lead to
criminal charges being brought against the mayor.
\\
\hline Reference & En conf\'erence de presse, jeudi, M. Blair a affirm\'e qu'il n'y avait rien dans cette vid\'eo qui puisse constituer des "motifs raisonnables" pouvant mener au d\'ep\^ot d'une accusation criminelle contre le maire.
\\
\hline RNNenc-50 & Lors de la conf\'erence de presse de jeudi, M. Blair a dit qu'il n'y avait rien dans cette vid\'eo qui pourrait constituer une "motivation raisonnable" pouvant entraîner des accusations criminelles port\'ees contre le maire.
\\
\hline RNNsearch-50 & Lors d'une conf\'erence de presse jeudi, M. Blair a d\'eclar\'e qu'il n'y avait rien dans cette vid\'eo qui pourrait constituer un "motif raisonnable" qui pourrait conduire à des accusations criminelles contre le maire.
\\
\hline Google \mbox{Translate} & 
Lors d'une conf\'erence de presse jeudi, M. Blair a d\'eclar\'e qu'il n'y avait rien
dans cette vido qui pourrait constituer un "motif raisonnable" qui pourrait
mener à des accusations criminelles portes contre le maire.
\\
\hline
        \end{tabular}
    \end{minipage}
    \caption{The translations generated by RNNenc-50 and RNNsearch-50 from long
        source sentences (30 words or more) selected from the test set. For each
        source sentence, we also show the gold-standard translation.  The
        translations by Google Translate were made on 27 August 2014.
}

\label{tab:translations}
\end{table}